\documentclass[10pt,twocolumn,letterpaper]{article}

\usepackage[pagenumbers]{wacv} %

\usepackage{graphicx}
\usepackage{amsmath}
\usepackage{amssymb}
\usepackage{booktabs}

\usepackage[pagebackref,breaklinks,colorlinks]{hyperref}

\usepackage{multirow}

\usepackage[capitalize]{cleveref}
\crefname{section}{Sec.}{Secs.}
\Crefname{section}{Section}{Sections}
\Crefname{table}{Table}{Tables}
\crefname{table}{Tab.}{Tabs.}

\def\wacvPaperID{2050} %
\def\confName{WACV}
\def\confYear{2024}

\begin{document}

\title{Reading Between the Mud: A Challenging Motorcycle Racer Number Dataset}

\author{Jacob Tyo\\
Carnegie Mellon University\\
DEVCOM Army Research Laboratory\\
{\tt\small jtyo@cs.cmu.edu}
\and
Youngseog Chung\\
Carnegie Mellon University\\
{\tt\small young@cs.cmu.edu}
\and
Motolani Olarinre\\
Carnegie Mellon University\\
{\tt\small tolani@cs.cmu.edu}
\and
Zachary C. Lipton\\
Carnegie Mellon University\\
{\tt\small zlipton@cmu.edu}
}
\maketitle

\begin{abstract}
    This paper introduces the off-road motorcycle Racer number Dataset (RnD), 
    a new challenging dataset 
    for optical character recognition (OCR) research. 
    RnD contains 2,411 images 
    from professional motorsports photographers that  
    depict motorcycle racers in off-road competitions. 
    The images exhibit a wide variety of factors that make OCR difficult, 
    including mud occlusions, 
    motion blur, non-standard fonts, glare, complex backgrounds, etc. 
    The dataset has 5,578 manually annotated 
    bounding boxes around visible motorcycle numbers, 
    along with transcribed digits and letters. 
    Our experiments benchmark leading OCR algorithms and 
    reveal an end-to-end F1 score of only 0.527 on RnD, 
    even after fine-tuning. 
    Analysis of performance on different occlusion types 
    shows mud as the primary challenge, 
    degrading accuracy substantially compared to normal conditions. 
    But the models struggle with other factors 
    including glare, blur, shadows, and dust. 
    Analysis exposes substantial room for improvement
    and highlights failure cases of existing models. 
    RnD represents a valuable new benchmark 
    to drive innovation in real-world OCR capabilities. 
    The authors hope the community will 
    build upon this dataset and baseline experiments 
    to make progress on the open problem 
    of robustly recognizing text 
    in unconstrained natural environments.
    The dataset is available at \url{https://github.com/JacobTyo/SwinTextSpotter}.
\end{abstract}

\begin{figure}[h]
    \centering
    \begin{subfigure}[b]{0.5\textwidth}
        \centering
        \includegraphics[width=\textwidth]{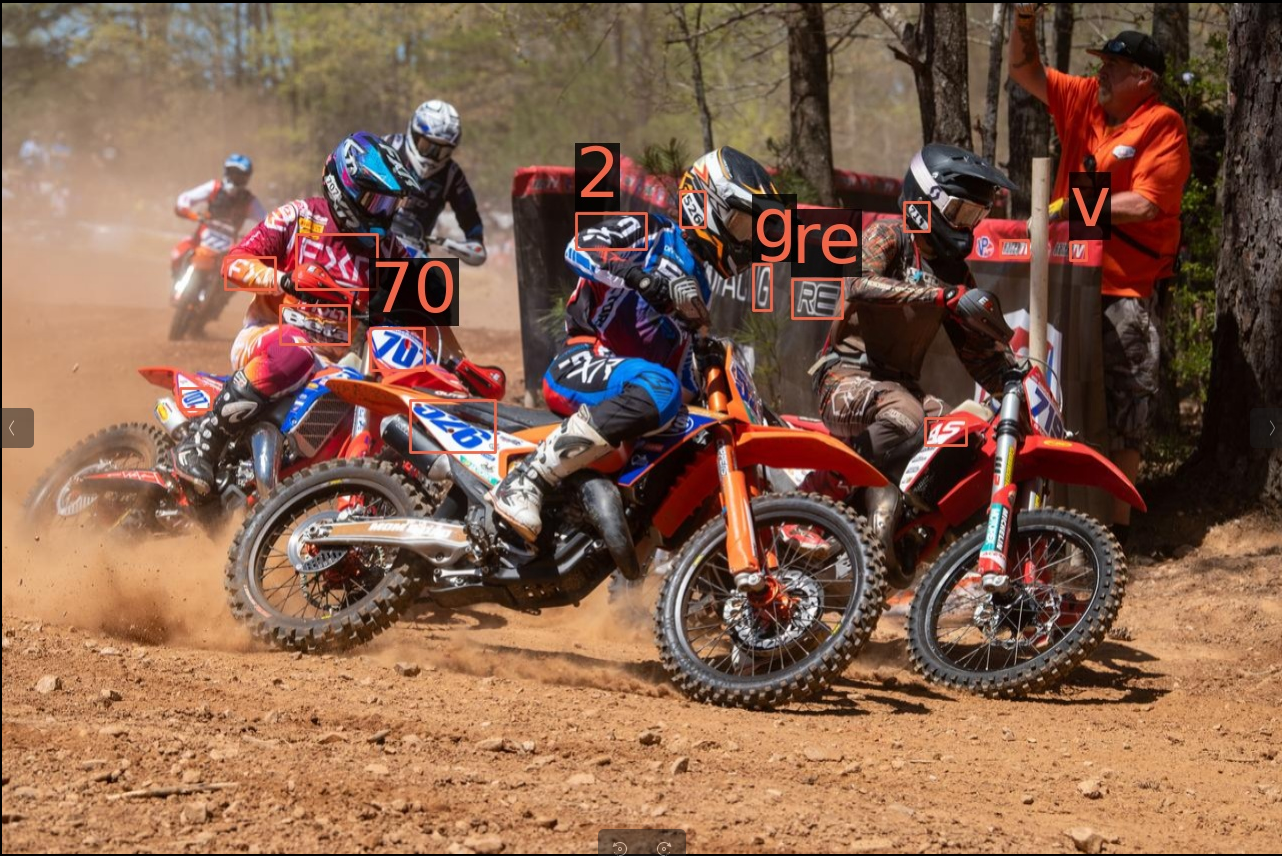}
    \end{subfigure}
    \hfill  %
    \begin{subfigure}[b]{0.5\textwidth}
        \centering
        \includegraphics[width=\textwidth]{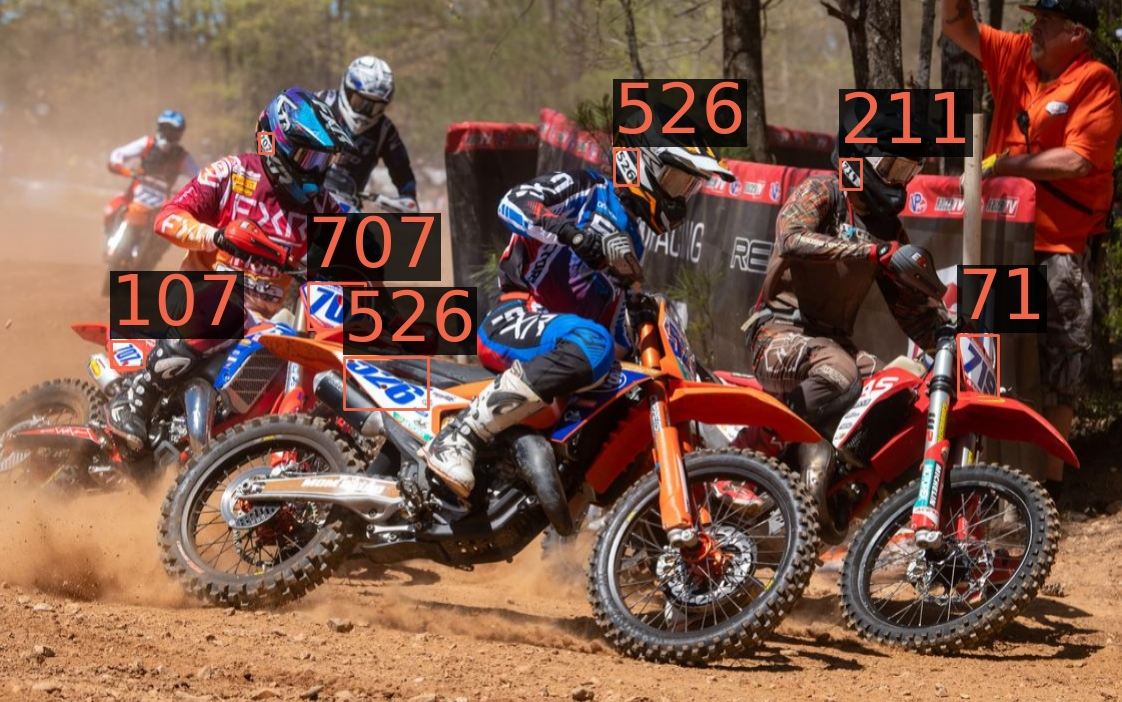}
    \end{subfigure}
    \caption{Detecting and recognizing numbers on motorcycles at the start of a race. The top image displays the detected text from a state-of-the-art off-the-shelf OCR model - many of the numbers are not detected or not recognized (bounding boxes with no text prediction). The bottom image displays the detected text from the same model which was further fine-tuned on RnD. 
    }
    \label{fig:headline}
\end{figure}

\section{Introduction}

Optical character recognition (OCR) 
is a well-studied task in computer vision with immense practical utility. 
There are many widely deployed systems that 
require detecting and recognizing textual information from visual data.
Thanks to developments in deep learning techniques combined with large annotated datasets, 
models can now accurately detect and recognize text 
in images across many languages, contexts and visual domains. 
Throughout much of its development,
research and datasets in OCR 
have focused on standardized fonts 
in structured environments, 
such as typed documents, 
road signs, and license plates, 
and OCR systems developed under such controlled conditions 
are fairly robust and can produce accurate predictions within their corresponding 
domain~\cite{Appalaraju_2021_ICCV, shashirangana2020automated, netzer2011reading}.  

A much more challenging, 
but much more versatile setting is recognizing text 
in unstructured and natural settings. 
However, recognizing text ``in the wild'' 
with unconstrained fonts, 
orientations, layouts, and contexts 
remains an open challenge~\cite{textinthewild}. 
While it is possible to steer the OCR system 
to be more robust towards particular settings 
(e.g. poor lighting)  %
by collecting and annotating data exposed to such conditions, 
in reality, a natural scene could present a myriad 
of diverse conditions which can undermine 
the system's ability to produce accurate text predictions.   
Furthermore, new domains emerge where current OCR methods struggle 
due to unique factors previously unseen in existing datasets.

One domain that presents a wide variety of challenging conditions for OCR is recognizing the racer numbers on motorcycles 
and all-terrain vehicles (ATVs) 
during off-road racing events 
(collectively referred to as \emph{motorcycles} 
in this paper).
Racer numbers, which can be used to identify the racer, are affixed on various locations of each racer and their vehicle. 
Accurate OCR for racer numbers can enable various useful applications, such as tracking race standings and automated analytics.
However, due to the off-road nature of these events, 
the numbers inevitably exhibit 
a combination of mud occlusions, 
non-standard layouts,
complex backgrounds, 
glare, 
and heavy motion blur.
Each of these conditions in isolation presents a major challenge for OCR, and their combination makes this an even more difficult task.
Further, to the authors' best knowledge, there exists 
no public dataset which can support research to tackle these challenges.

To address this gap,
we introduce the off-road motorcycle Racer number Dataset (RnD). 
RnD contains 2,411 images sampled 
from 16 professional motorsports photographers 
across 50 different off-road events. 
The images exhibit the unique challenges 
of this domain: 
mud covering numbers, 
scratches and dirt obfuscating digits, 
heavy shadows and glare from uncontrolled outdoor lighting, 
complex backgrounds of other vehicles, 
bystanders, trees, and terrain, 
motion blur from rapid maneuvers, 
large variations in racer number size and location on motorcycles, 
and various fonts and colors chosen by each racer.

The images are annotated with polygons around 
every visible motorcycle number 
along with the transcribed sequence of digits and letters.
Only racer identifying texts were annotated.
The images were sourced from real racing competitions which span
diverse track conditions, 
weather, lighting, bike types, and racer gear.

The rest of this paper is structured as follows.
We first discuss the dataset contents and highlight the domain gaps from existing OCR datasets. 
We detail the annotation protocol tailored to this domain. 
We then benchmark leading OCR algorithms 
to establish baseline accuracy on RnD. 
The experiments reveal substantial room for improvement, 
which motivates further research into techniques that can robustly handle
mud occlusion, and rapidly evolving perspectives. 
Our dataset provides the imagery 
to support developing and evaluating such advances in OCR.

The main contributions are:
\begin{itemize}
    \item RnD: 
    a off-road motorcycle Racer number Dataset containing 
    2,411 images with 5,578 labeled numbers 
    sampled from professional photographers 
    at 50 distinct off-road races. 
    To our knowledge, 
    this is the first large-scale dataset 
    focused on recognizing racer numbers in
    off-road motorsports imagery.
    \item A rigorous benchmark 
    of generic state-of-the-art OCR models, 
    revealing poor accuracy on RnD 
    and substantial room for innovation.
    \item Experiments comparing off-the-shelf
    and fine-tuning strategies. 
    Even the best fine-tuned models fall short.
    \item Qualitative analysis of prediction errors which provides insights 
    into failure modes to guide future research directions.
\end{itemize}

We hope RnD and our initial experiments 
will catalyze innovation 
in real-world text recognition capabilities. 
Robust reading of racer numbers 
has potential applications in race analytics, 
timing systems, 
media broadcasts, 
and more. 
Our work reveals this as an open research problem 
necessitating domain-targeted techniques.

\section{Related Work}

Text detection and recognition in images 
is a classic computer vision task. 
Early traditional methods relied on sliding windows, 
connected components, 
and handcrafted features like HOG~\cite{wang2010word}. 
With the advent of deep learning, 
convolutional and recurrent neural networks 
now dominate scene text recognition pipelines~\cite{textinthewild}. 
Models leverage large annotated datasets 
to learn powerful representations tuned for 
text detection and recognition in a specific domain.

Many datasets and competitions have driven progress 
in general OCR. 
These include ICDAR~\cite{karatzas2013icdar}, 
COCO-Text~\cite{lin2014microsoft}, 
and Street View Text~\cite{wang2011end}.
Popular detection models build on 
Region Proposal Networks 
and include CTPN~\cite{tian2016detecting}, 
EAST~\cite{Zhou_2017_CVPR}, 
and Craft~\cite{baek2019character}. 
Recognition is often achieved via CNN + RNN architectures 
like CRNN~\cite{CRNN} 
or transformer networks like ASTER~\cite{aster}. 
More recent state-of-the-art methods utilize 
pre-trained vision models like 
ViTSTR~\cite{vitstr}, 
PARSeq~\cite{bautista2022parseq}, 
CLIP4STR~\cite{zhao2023clip4str},
and DeepSolo~\cite{Ye_2023_CVPR}. 
However, 
most OCR research targets 
images of documents, 
signs, or web images. 
While many of these works aim to go beyond structured settings 
(e.g.images of documents, signs, or web images)
and address the task of ``robust reading'', i.e. OCR in incidental or real scenes,
recognizing text ``in the wild''
with few assumptions remains an open challenge~\cite{textinthewild}.
Furthermore, 
domain gaps exist where current methods fail 
on specialized applications. 
Our work focuses on one such gap - 
recognizing racer numbers in motorsports.

\begin{figure*}[h]
    \centering
    \begin{subfigure}[b]{0.25\textwidth}
        \centering
        \includegraphics[width=\textwidth]{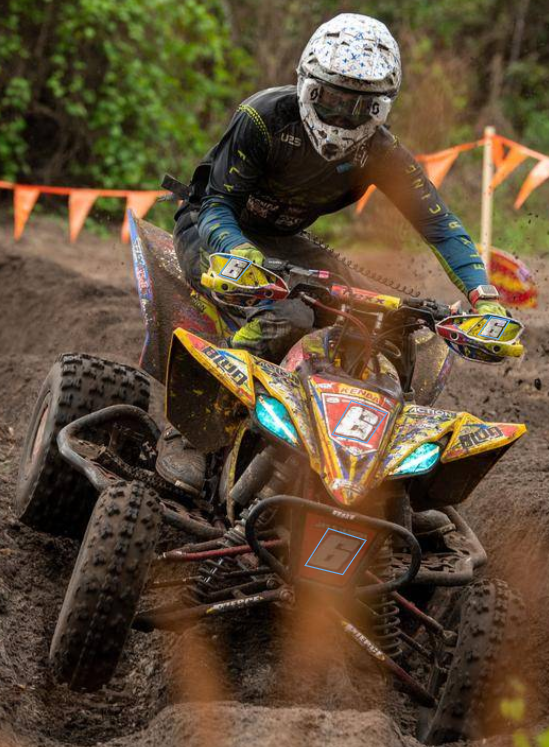}
        \caption{}
    \end{subfigure}
    \hfill  %
    \begin{subfigure}[b]{0.211\textwidth}
        \centering
        \includegraphics[width=\textwidth]{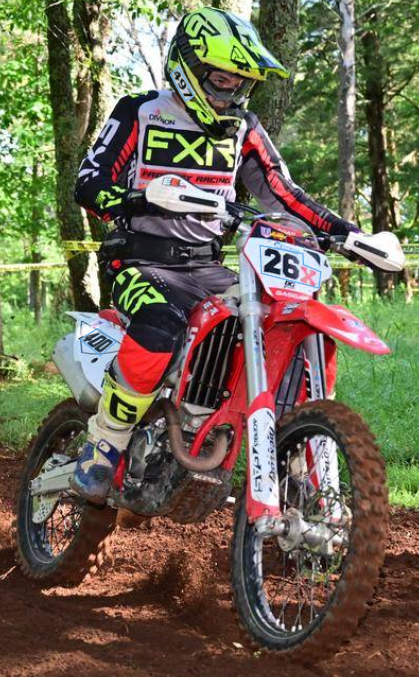}
        \caption{}
    \end{subfigure}
    \hfill
    \begin{subfigure}[b]{0.25\textwidth}
        \centering
        \includegraphics[width=\textwidth]{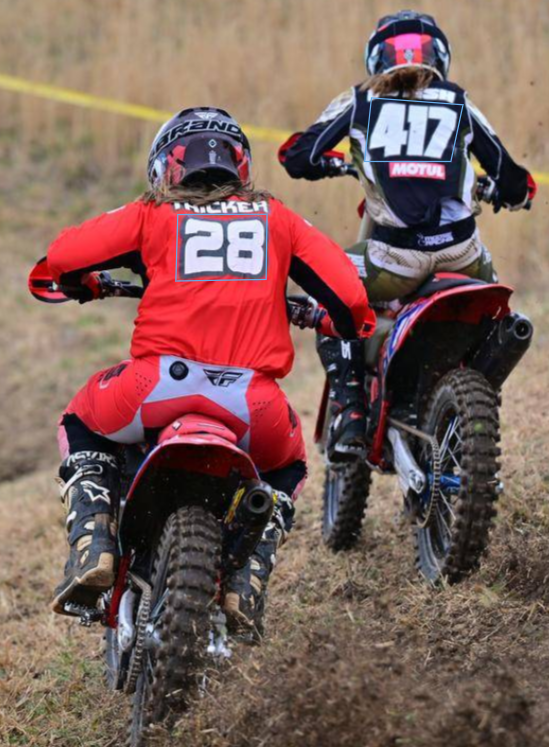}
        \caption{}
    \end{subfigure}
    \hfill
    \begin{subfigure}[b]{0.266\textwidth}
        \centering
        \includegraphics[width=\textwidth]{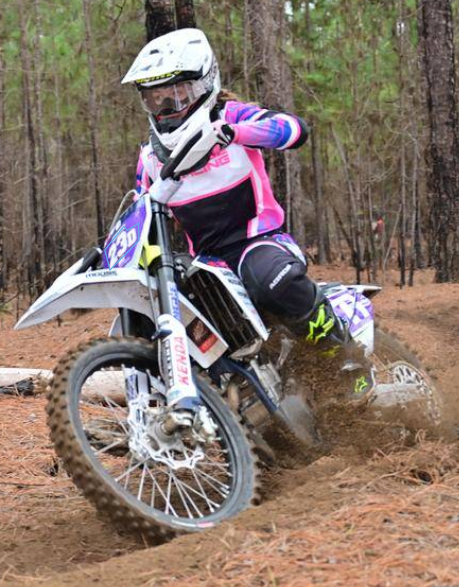}
        \caption{}
    \end{subfigure}
    \caption{Common locations and variations of racer numbers. 
    (a) Numbers can be seen on the hand guards, and vegetation close to the photographer makes for a new sort of occlusions. 
    (b) The front number, side number, and helmet number are all different. 
    (c) Numbers can be on the back of racer's jerseys.
    (d) Different front and side numbers.
    }  
    \label{fig:how_labeled}
\end{figure*}

A few prior works address detecting 
and recognizing 
the license plates on vehicles~\cite{ap2020automatic, laroca2018robust, chen2019automatic, lee2019snider, silva2018license, quang2022character, laroca2021efficient}.
Some have focused specifically on street motorcycle 
number plates~\cite{kulkarni2018automatic, sathe2022helmet, sanjana2021review, lee2004extraction}.
All of these efforts use data gathered from some form of 
street camera, 
which are placed in strategic locations 
with recognizing license plates specifically in mind. 
In contrast, 
our dataset is gathered from 
professional motorsport photographers 
focused on capturing the most aesthetically pleasing 
photograph of each racer. 
Furthermore, 
existing datasets have standardized plates 
which differ greatly 
from the diverse layouts and occlusions 
of off-road motorcycle numbers. 
Street motorcycle plates exhibit 
consistency in position and appearance, 
unlike the numbers encountered during off-road competitions. 
The conditions during races also introduce 
and exacerbate factors like motion blur, 
mud occlusion, 
glare, 
and shaky cameras not prevalent in street imagery. 
RnD provides novel real-world imagery to push OCR capabilities.

The most relevant prior domain 
is recognizing runner bib numbers 
in marathon images~\cite{shivakumara2017new, ben2012racing, boonsim2018racing, kamlesh2017person}. 
This shares similarities, 
but runner bibs provide 
more spatial and appearance consistency 
than motorcycle racing numbers. 
Datasets like TGCRBNW~\cite{TGCRBNW} 
exhibit some motion blur and night racing, 
but do not contain the mud, 
vehicle occlusion and diversity of layouts seen in motorsports.

Number recognition has also been studied in other sports - 
football~\cite{yamamoto2013multiple, bhargavi2022knock}, 
soccer~\cite{Gerke_2015_ICCV_Workshops, gerke2017soccer, vsaric2008player, diop2022soccer, alhejaily2023automatic}, 
basketball~\cite{ahammed2018basketball}, 
track and 
field~\cite{messelodi2013scene}, 
and more~\cite{liu2019pose, nag2019crnn, vats2021multi, wronska2017athlete}.
However, most focus on jersey numbers in 
commercial broadcast footage rather 
than track/field-side imagery. 
Existing sports datasets offer limited diversity and size. 
To our knowledge, 
RnD represents the largest, 
most varied collection of motorsports numbers in natural contexts.

In summary, 
prior work has made great progress in OCR 
for documents, signs, and other domains,
but real-world applications like 
recognizing racers in off-road competitions 
remain extremely challenging due to domain gaps in current data. 
RnD provides novel imagery to spur advances in OCR for motorsports. 
Our benchmark experiments expose substantial room 
for improvement using this data.

\begin{figure*}[h]
    \centering
    \begin{subfigure}[b]{0.1905\textwidth}
        \centering
        \includegraphics[width=\textwidth]{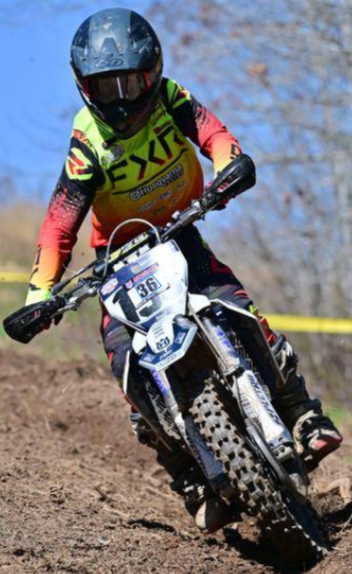}
        \caption{}
        \label{img:overlap_glare}
    \end{subfigure}
    \hfill  %
    \begin{subfigure}[b]{0.1575\textwidth}
        \centering
        \includegraphics[width=\textwidth]{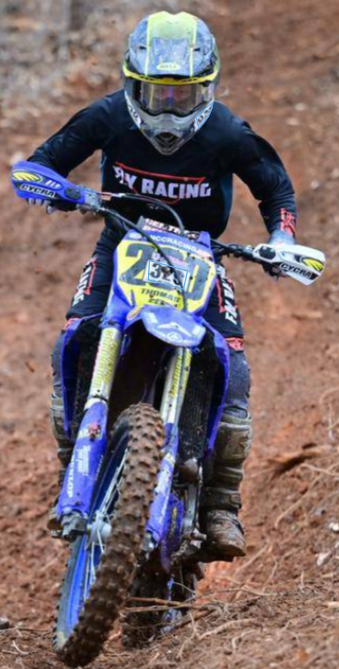}
        \caption{}
        \label{img:overlap_brake}
    \end{subfigure}
    \hfill
    \begin{subfigure}[b]{0.325\textwidth}
        \centering
        \includegraphics[width=\textwidth]{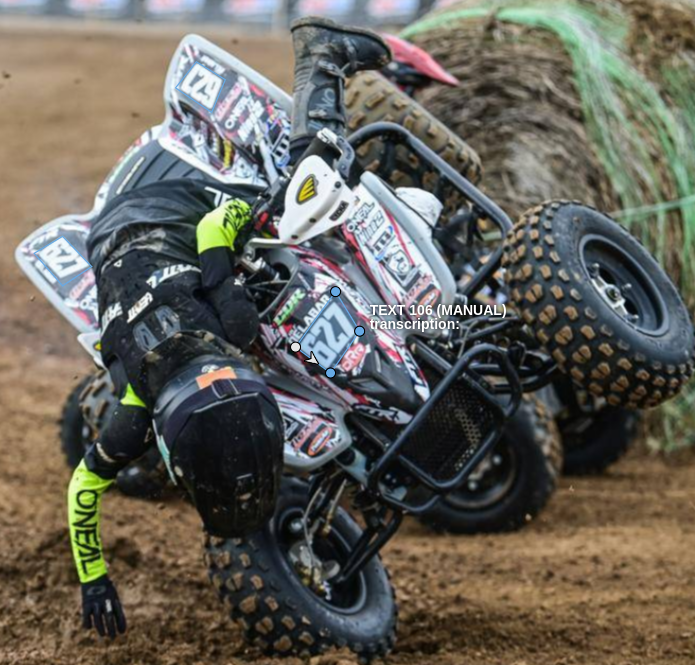}
        \caption{}
        \label{img:atv_crash}
    \end{subfigure}
    \hfill
    \begin{subfigure}[b]{0.311\textwidth}
        \centering
        \includegraphics[width=\textwidth]{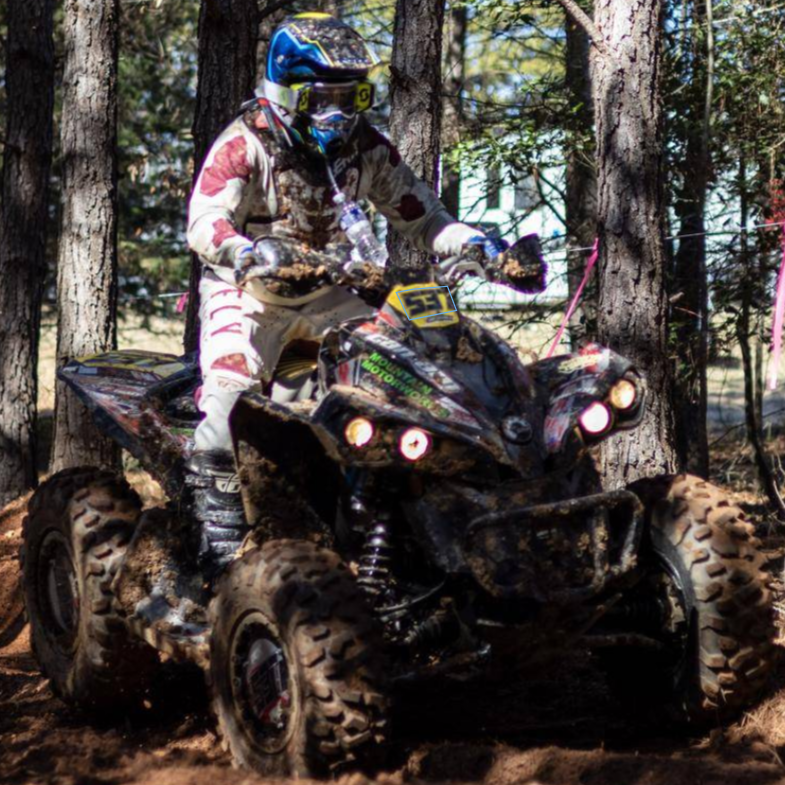}
        \caption{}
        \label{img:shadows}
    \end{subfigure}
    \caption{Examples of some difficult, but not muddy, images. 
    (a) Two separate numbers are on the front of the motorcycle, a smaller number overlapping a bigger number. Furthermore, half of the number plate is not legible due to glare.
    (b) The front-brake cable overlaps the number.
    (c) A racer is crashing, resulting in contrived number orientations.
    (d) Shadows cast from trees cause difficult lighting conditions.}  
    \label{fig:overlapping}
\end{figure*}

\begin{figure*}[h]
    \centering
    \begin{subfigure}[b]{0.161\textwidth}
        \centering
        \includegraphics[width=\textwidth]{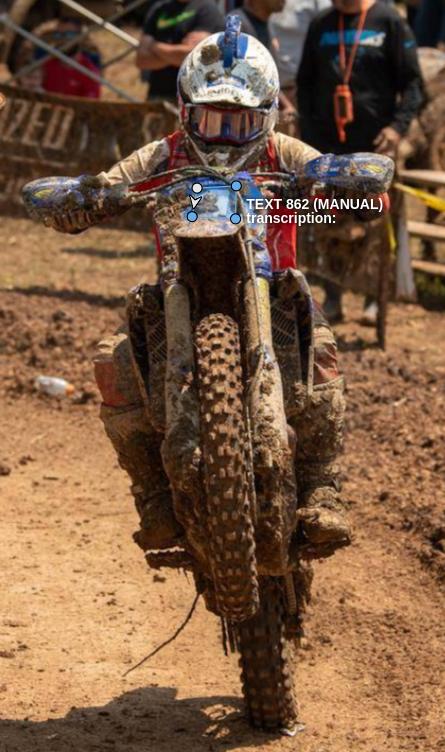}
        \caption{}
    \end{subfigure}
    \hfill  %
    \begin{subfigure}[b]{0.265\textwidth}
        \centering
        \includegraphics[width=\textwidth]{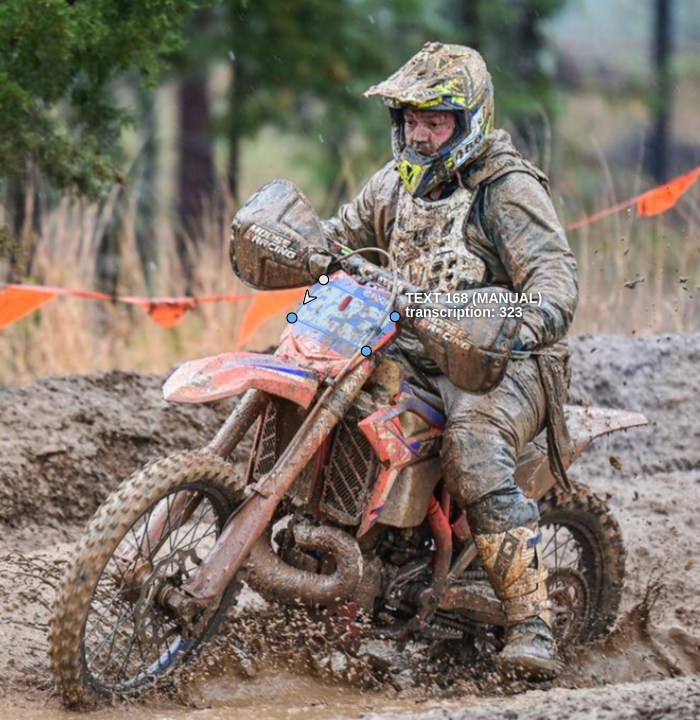}
        \caption{}
    \end{subfigure}
    \hfill
    \begin{subfigure}[b]{0.265\textwidth}
        \centering
        \includegraphics[width=\textwidth]{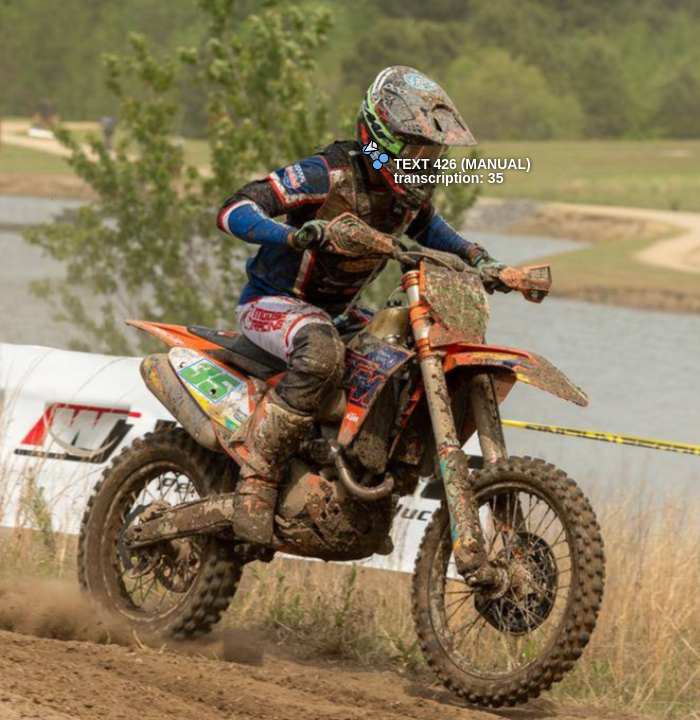}
        \caption{}
    \end{subfigure}
    \hfill
    \begin{subfigure}[b]{0.146\textwidth}
        \centering
        \includegraphics[width=\textwidth]{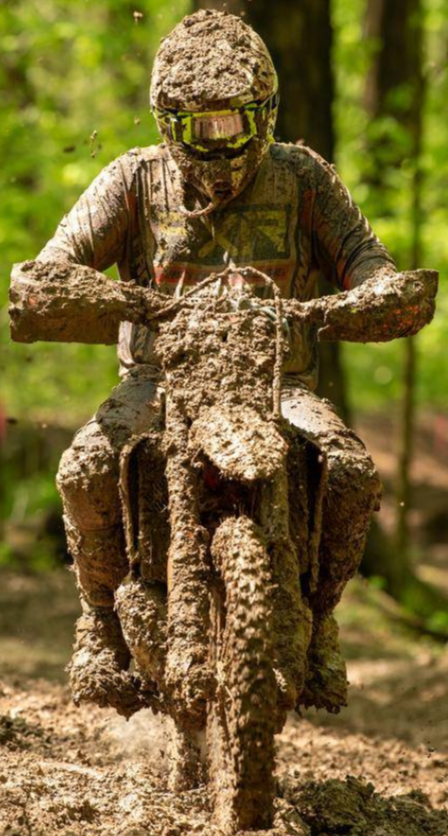}
        \caption{}
        \label{img:completemud}
    \end{subfigure}
    \hfill
    \begin{subfigure}[b]{0.1375\textwidth}
        \centering
        \includegraphics[width=\textwidth]{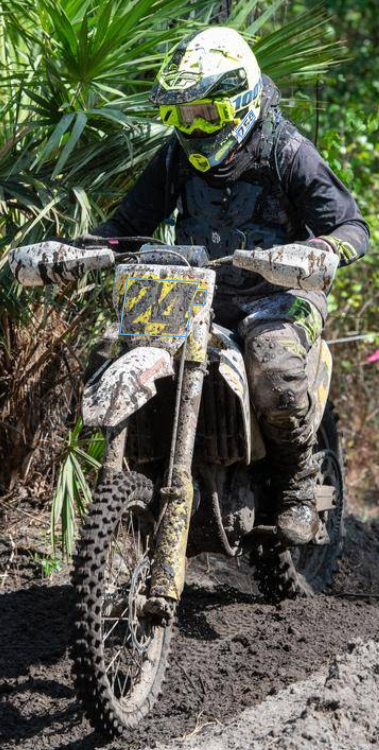}
        \caption{}
    \end{subfigure}
    \caption{Mud poses the most significant challenge to effective OCR in this domain. (a) Not only is the racer in an odd pose, but the number is also occluded in sticky mud. (b) The racer is covered in wet mud, posing a different, although more managable, type of mud occlusion. (c) Mud occlusions in sandy environments again poses new types of occlusions. (d) An extreme example of sticky mud completely obscuring all details about a racers number. (e) Generic example of the most commonly seen type of mud occlusion.}  
    \label{fig:mud}
\end{figure*}

\section{Dataset}

The off-road motorcycle Racer number Dataset (RnD)\footnote{
The dataset is available at \url{https://github.com/JacobTyo/SwinTextSpotter}.}
is comprised of 2,411
images gathered from
the off-road photography platform
\url{PerformancePhoto.co}. 
Each image depicts motorcycle racers engaged in competitive events, 
with visible racer numbers on themselves and their motorcycles.
The dataset includes bounding box annotations 
and transcriptions from over 50 different off-road 
motorcycle and ATV races. 
The races cover various track conditions, 
weather, 
and lighting. 
The images were captured by 16 different professional photographers using a wide range of high-end cameras.

Racers can have anywhere from one to as many as 20 numbers 
located on their body and motorcycle. 
The common locations for a number include
the front and sides of the motorcycle,
on the cheeks of the racer's helmet,
and on the back of the racers jersey. 
However, in rare cases, 
numbers can also be seen on the wheels 
and handguards. 
The numbers on a single racer and vehicle
do not need to all be the same number. 
Commonly, 
the numbers on the helmet do not match the numbers on 
the motorcycle, 
and the number on the front of the motorcycle
does not need to match the number on the side.
It is also common for numbers to only be present on the racer, 
but not on the motorcycle.
Figure~\ref{fig:how_labeled} highlights some of these examples. 

In RnD, there is a total of 5,578 racer number annotations.
The numbers can span from 1 to 5 characters in length, 
optionally including alphabetical characters
(e.g., adding a letter to the end of a number is 
a common modifier - 
for convenience, we still refer to all of these as \emph{numbers}). 
6\% of the dataset includes numbers that have 
alphabetical characters in addition to the numerics. 
The dataset is split randomly into a training and a testing set, 
with 80\% of the images in the training set. 

\begin{table*}[t]
\centering
\caption{Comparison of the text detection and recognition performance on the RnD test set using off-the-shelf versus fine-tuned state-of-the-art OCR models. Precision, recall, and F1 score are reported for both detection (Det-P, Det-R, Det-F1) and end-to-end recognition (E2E-P, E2E-R, E2E-F1). The off-the-shelf versions achieve very low scores, while fine-tuning improves results substantially. However, even fine-tuned models fall short of real-world viability, with the best YAMTS model obtaining only 0.527 end-to-end F1 score. This highlights significant room for improvement using domain-targeted techniques and data such as RnD.}
\label{tab:results}
\begin{tabular}{@{}cccccccc@{}}
\toprule
\multicolumn{2}{c}{Model}                         & Det-P & Det-R & Det-F1 & E2E-P & E2E-R & E2E-F1 \\ \midrule
\multirow{2}{*}{Off-the-shelf}        & SwinTS &   0.195    &  0.287     &  0.2323      &  0.101     & 0.148      & 0.120       \\
                                      & YAMTS   &  0.192     &  0.491     &  0.276      &   0.106    &  0.244     &  0.148      \\ \midrule
\multirow{2}{*}{Fine-Tuned}           & SwinTS &  0.810     &  0.673     &  0.734      &   0.513    & 0.415      &   0.459     \\
                                      & YAMTS   &  0.847     &  0.715     &  0.775      &    0.758   & 0.404      &  \textbf{0.527}      \\ \bottomrule
\end{tabular}
\end{table*}

\subsection{Annotation Process}

Only the racer numbers were annotated
instead of all visible text by one of the authors.
All visible racer numbers were tightly bounded by a polygon
(i.e. the bounding box),
and each polygon is tagged with the characters contained within 
(i.e. the number).
If a character was ambiguous or unclear, 
it was labeled with a `\#' symbol. 
Only 
the humanly identifiable text was transcribed. 
Any racer numbers that were fully occluded 
or too blurry to discern were not annotated.

The transcription task was restricted 
to only use the context of each 
individual bounded region. 
The full image context could not be used 
to infer ambiguous numbers based on 
other instances of that racer's number 
elsewhere on the motorcycle. 
This simulates the local context available 
to optical character recognition models.

\subsection{Analysis}

Figure~\ref{fig:overlapping} highlights some of the challenging factors present in this dataset. 
Lighting conditions vary from extremely bright to extremely dark (including night races).
Figure~\ref{img:overlap_glare} gives an example of glare 
that is common in a field with exposure to sunlight (8\% of images),  
and Figure~\ref{img:shadows} shows the complications that 
the forest can cause on lighting conditions (7\% of images).
Not only are there occlusions typical of other datasets
such as trees or other racers blocking the view,
but we are also presented with extremely challenging cases where a smaller 
number is placed over top of a bigger number (See Figure~\ref{img:overlap_glare}). 
In such cases, 
we label every number we can properly identify. 
Furthermore, 
as shown by the front brake cable in 
Figure~\ref{img:overlap_brake},
some motorcycles have components that pass in front of the 
number plate. 
Finally, orientation of the numbers vary greatly,  
not only due to the nature of motorcycles 
(i.e. they must be leaned over to turn corners), 
but also in cases such as crashes, 
as shown in Figure~\ref{img:atv_crash}.

The most unique aspect of this dataset is a 
new type of occlusion: mud. 
Mud is frequently encountered in off-road racing, 
and Figure~\ref{fig:mud} gives examples ranging from 
light to extreme (44\% of images). 
In the worst of cases, 
it is impossible to detect any racer numbers (Figure~\ref{img:completemud}). 
However, in many cases, 
humans are still able to accurately complete this task.

\section{Experiments}

We conducted experiments 
to benchmark the performance of modern OCR methods 
on the RnD. 
Our goals here are twofold: 
1) establish baseline results on this new domain, 
and 2) analyze where current algorithms fail.
Four NVIDIA Tesla V100 GPUs were used for these experiments.
Hyperparmeter searching was performed 

\subsection{Models}

Our experiments leverage two state-of-the-art scene text spotting models: 
\begin{itemize}
    \item \textbf{YAMTS}: Yet Another Mask Text Spotter \cite{krylov2021open}

     YAMTS is a Mask R-CNN-based model with an additional recognition head
    for end-to-end scene text spotting.
    A ResNet-50~\cite{he2016deep} is used for text detection, 
    with a convolutional text encoder and a GRU decoder. 
    \item \textbf{SwinTS}: Swin Text Spotter \cite{huang2022swintextspotter}
    
    The Swin Text Spotter is an end-to-end 
    Transformer-based model that improves detection 
    and recognition synergy through a recognition conversion module.
    A feature pyramid network is used to decrease the sensitivity to text size, 
    and the recognition conversion model 
    enables joint optimization of the detection 
    and recognition losses. 
\end{itemize}

\begin{figure}
    \centering
    \begin{subfigure}[b]{0.5\textwidth}
        \centering
        \includegraphics[width=\textwidth]{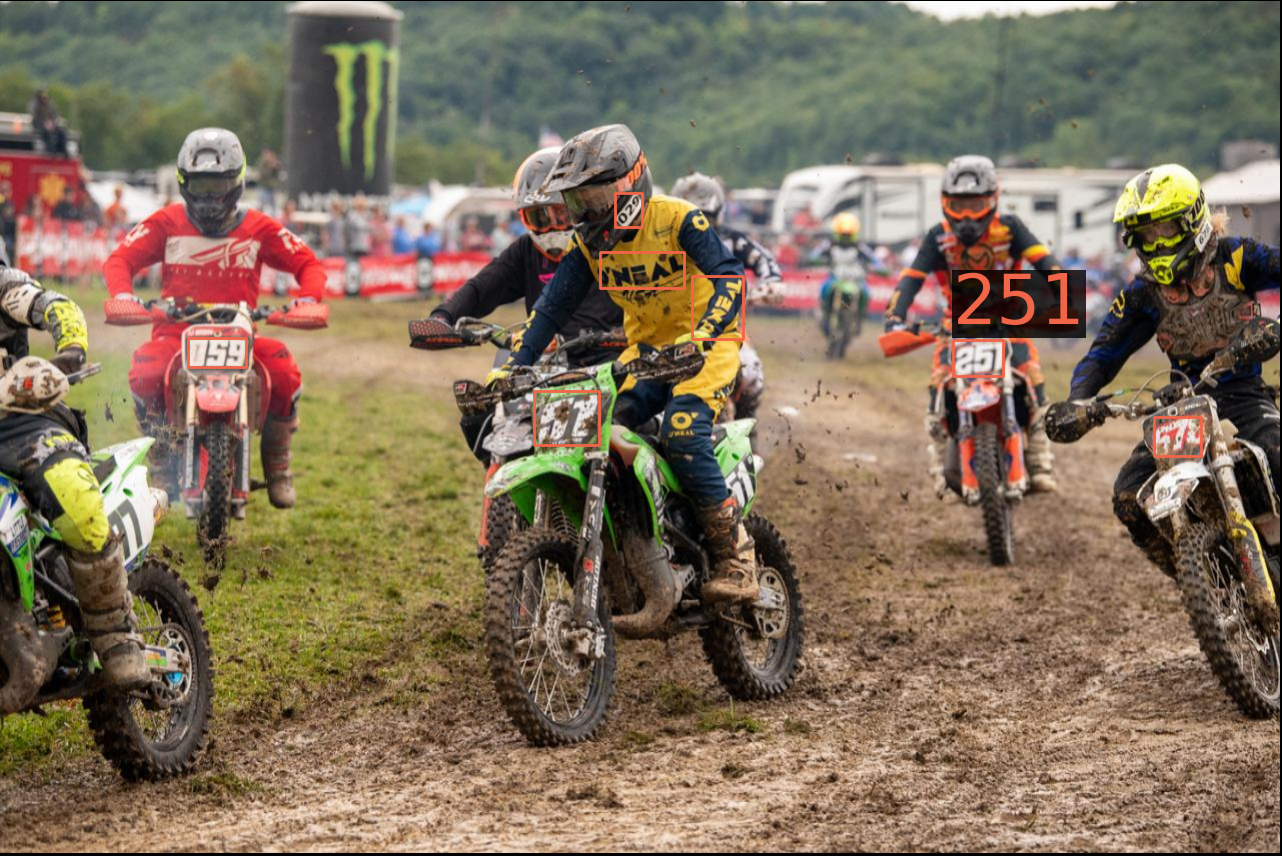}
    \end{subfigure}
    \hfill  %
    \begin{subfigure}[b]{0.5\textwidth}
        \centering
        \includegraphics[width=\textwidth]{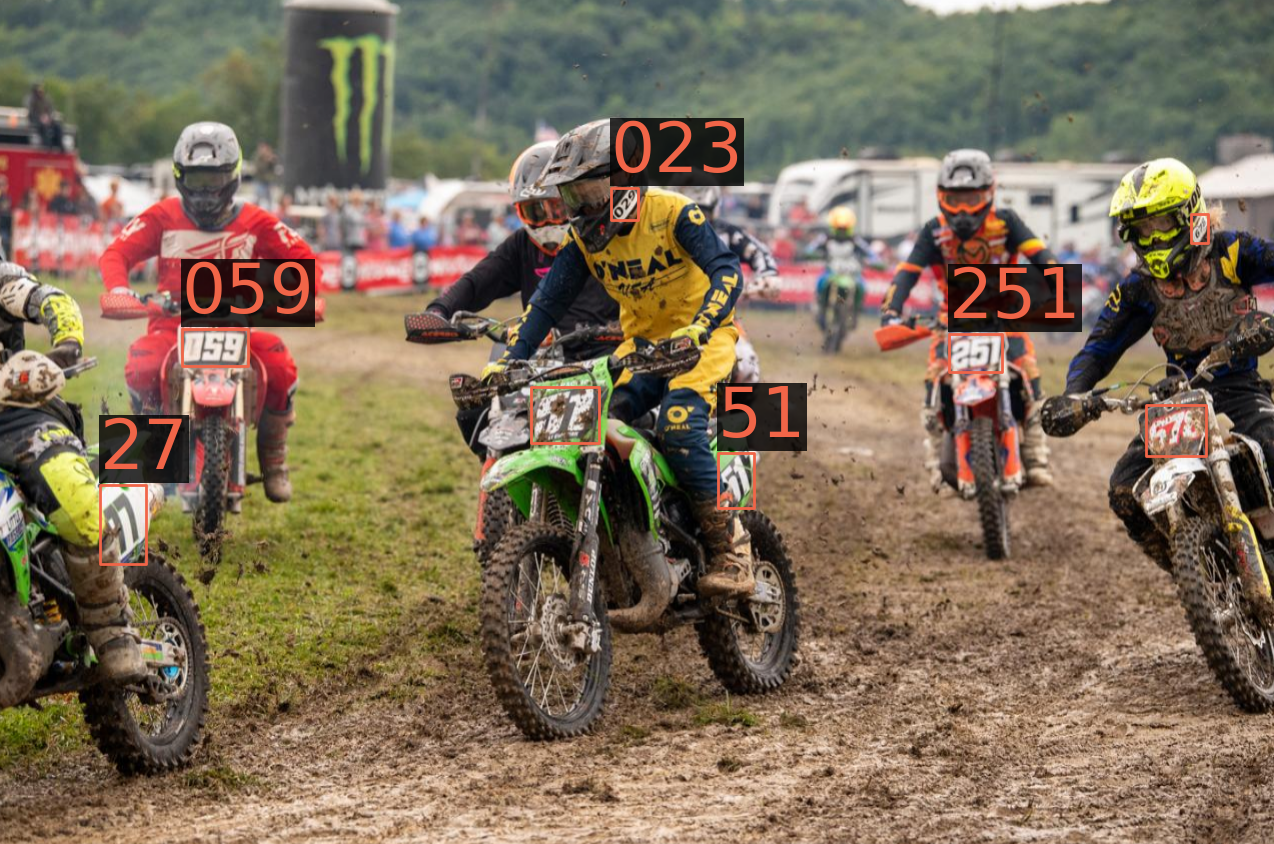}
    \end{subfigure}
    \caption{Example showcasing model successes and failures on a complex muddy image. The top image shows detected text from the off-the-shelf YAMTS model before fine-tuning, which recognizes only 1 number correctly (``251"). The bottom image displays results from the fine-tuned YAMTS model, which detects all 8 visible numbers but only correctly recognizes 3 of them. This highlights benefits of domain-specific fine-tuning, as the pre-trained model struggles. However, even the fine-tuned model has difficulty accurately recognizing highly degraded text, exposing substantial room for improvement.}
    \label{fig:muddy-start}
\end{figure}

\begin{figure}
    \centering
    \begin{subfigure}[b]{0.235\textwidth}
        \centering
        \includegraphics[width=\textwidth]{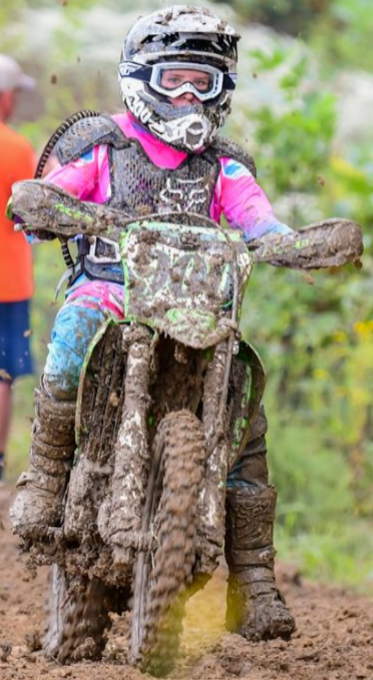}
    \end{subfigure}
    \hfill  %
    \begin{subfigure}[b]{0.235\textwidth}
        \centering
        \includegraphics[width=\textwidth]{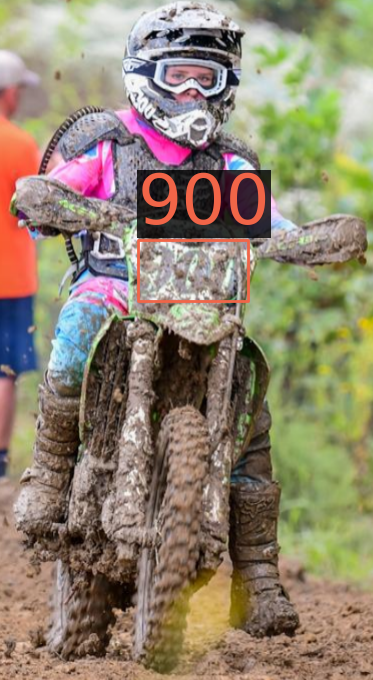}
    \end{subfigure}
    \caption{Example showcasing the fine-tuned model learning to see through mud. The left image depicts the predictions from the off-the-shelf YAMTS model before fine-tuning, which does not recognize any text. The right image displays results from the fine-tuned YAMTS model, which is able to see through the heavy mud occlusion and properly detect and recognize the racer number. This demonstrates improved robustness to real-world mud occlusion after domain-specific fine-tuning.}
    \label{fig:see-through-mud}
\end{figure}

\begin{figure*}[h]
    \centering
    \begin{subfigure}[b]{0.13\textwidth}
        \centering
        \includegraphics[width=\textwidth]{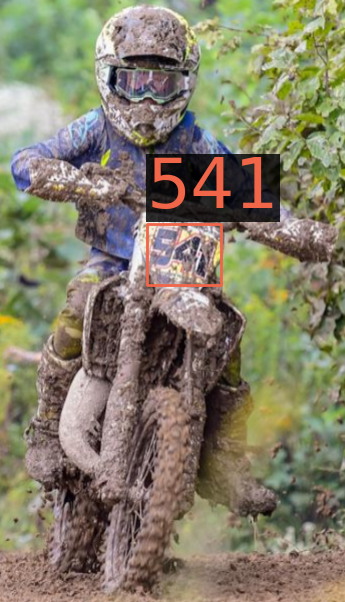}
        \caption{}
        \label{img:rtm2}
    \end{subfigure}
    \hfill  %
    \begin{subfigure}[b]{0.262\textwidth}
        \centering
        \includegraphics[width=\textwidth]{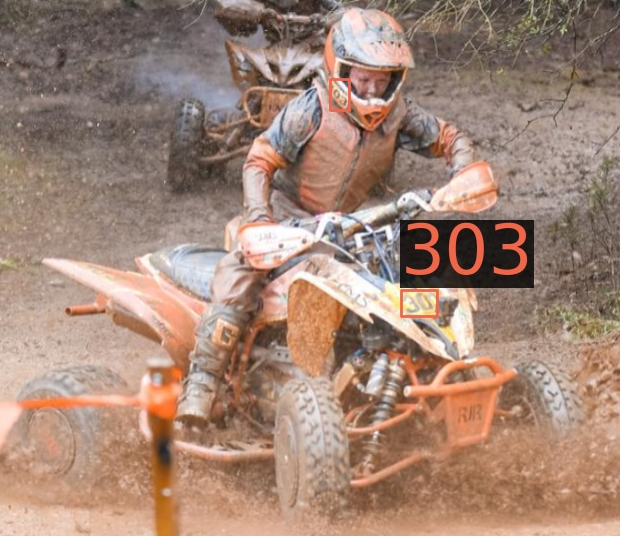}
        \caption{}
        \label{img:rtm1}
    \end{subfigure}
    \hfill
    \begin{subfigure}[b]{0.163\textwidth}
        \centering
        \includegraphics[width=\textwidth]{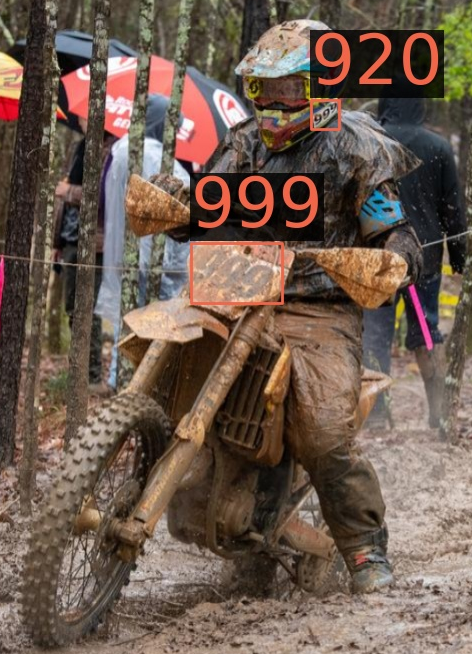}
        \caption{}
        \label{img:mudfail}
    \end{subfigure}
    \hfill
    \begin{subfigure}[b]{0.22\textwidth}
        \centering
        \includegraphics[width=\textwidth]{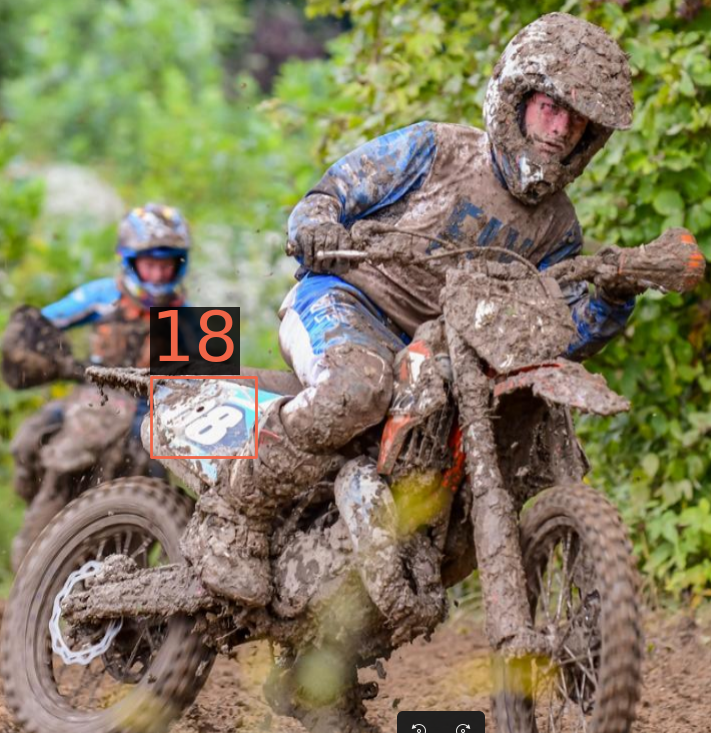}
        \caption{}
        \label{img:mudslidefail}
    \end{subfigure}
    \hfill
    \begin{subfigure}[b]{0.202\textwidth}
        \centering
        \includegraphics[width=\textwidth]{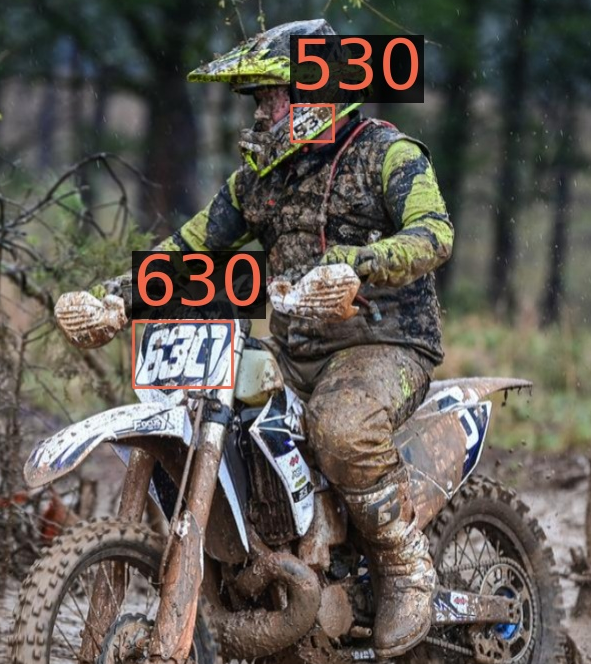}
        \caption{}
        \label{img:muddydude}
    \end{subfigure}
    \caption{Analysis of model performance on mud occluded numbers. (a) Model correctly recognizes front number by ignoring mud. (b) Quad number is recognized but muddy helmet number is missed. (c) Front number is read but very muddy helmet number is missed. (d) Number is detected but misrecognized due to odd position. (e) Two numbers are correctly read but muddy side number is missed.}  
    \label{fig:mudfails}
\end{figure*}

\begin{figure*}[h]
    \centering
    \begin{subfigure}[b]{0.24\textwidth}
        \centering
        \includegraphics[width=\textwidth]{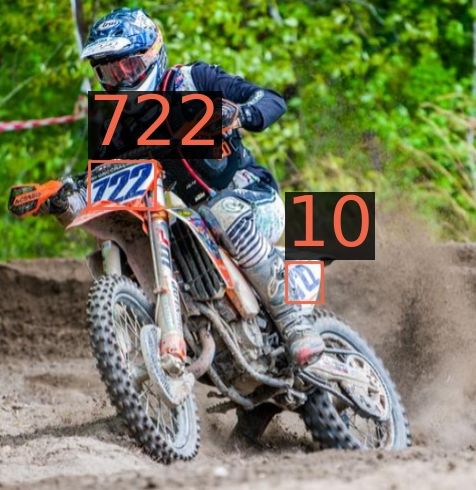}
        \caption{}
        \label{img:backbikefail}
    \end{subfigure}
    \hfill  %
    \begin{subfigure}[b]{0.165\textwidth}
        \centering
        \includegraphics[width=\textwidth]{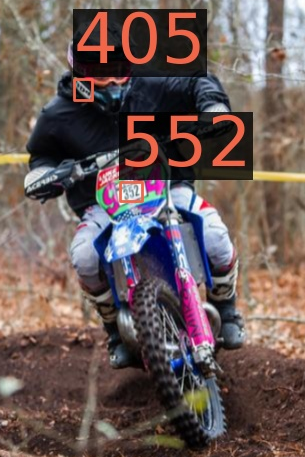}
        \caption{}
        \label{img:stackednums}
    \end{subfigure}
    \hfill
    \begin{subfigure}[b]{0.162\textwidth}
        \centering
        \includegraphics[width=\textwidth]{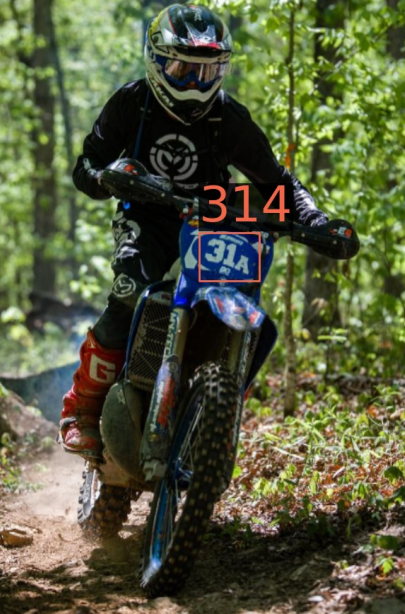}
        \caption{}
        \label{img:wrongletter}
    \end{subfigure}
    \hfill
    \begin{subfigure}[b]{0.115\textwidth}
        \centering
        \includegraphics[width=\textwidth]{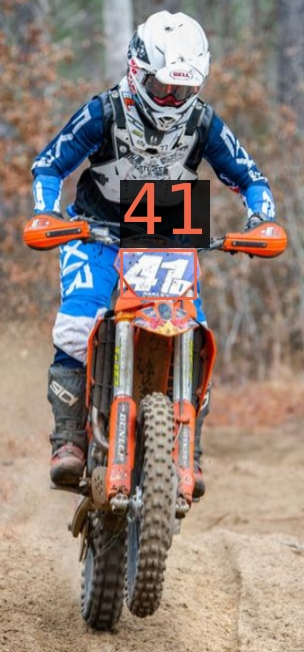}
        \caption{}
        \label{img:missedletter}
    \end{subfigure}
    \hfill
    \begin{subfigure}[b]{0.295\textwidth}
        \centering
        \includegraphics[width=\textwidth]{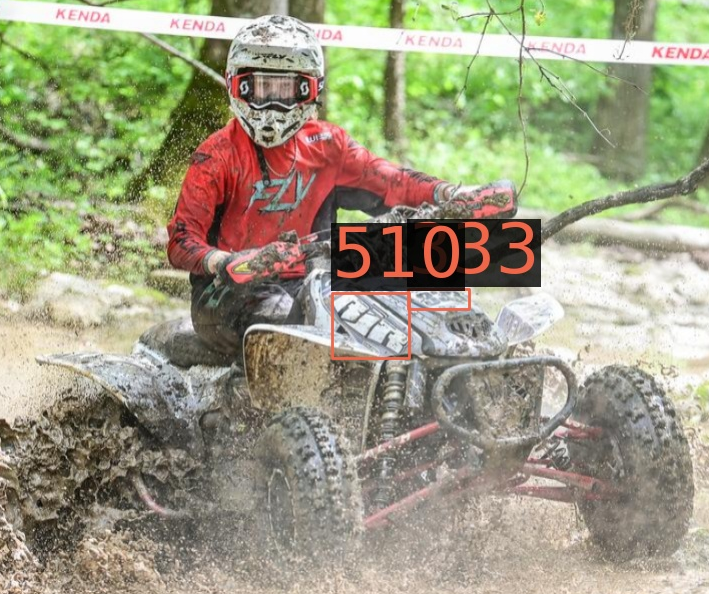}
        \caption{}
        \label{img:quadshard}
    \end{subfigure}
    \caption{Analysis of common non-mud failures: (a) Incorrect side number recognition. (b) Overlapping ``stacked'' numbers confuse the model. (c) A letter is mis-recognized as a number. (d) The letter portion of the racer number is missed. (e) Complex graphics on quad confuse model.}  
    \label{fig:cleanfails}
\end{figure*}

\begin{figure}
\centering
    \begin{subfigure}[b]{0.5\textwidth}
        \centering
        \includegraphics[width=\textwidth]{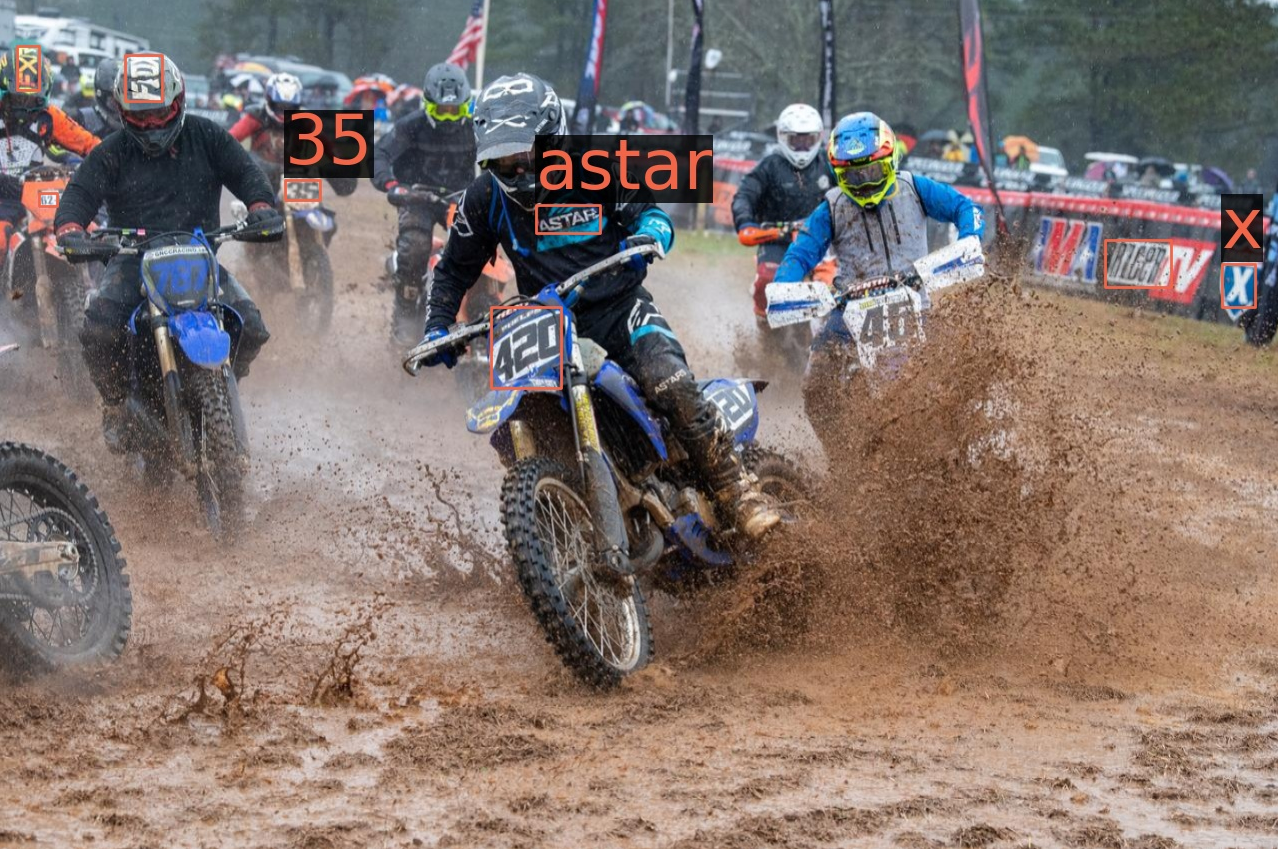}
    \end{subfigure}
    \hfill  %
    \begin{subfigure}[b]{0.5\textwidth}
        \centering
        \includegraphics[width=\textwidth]{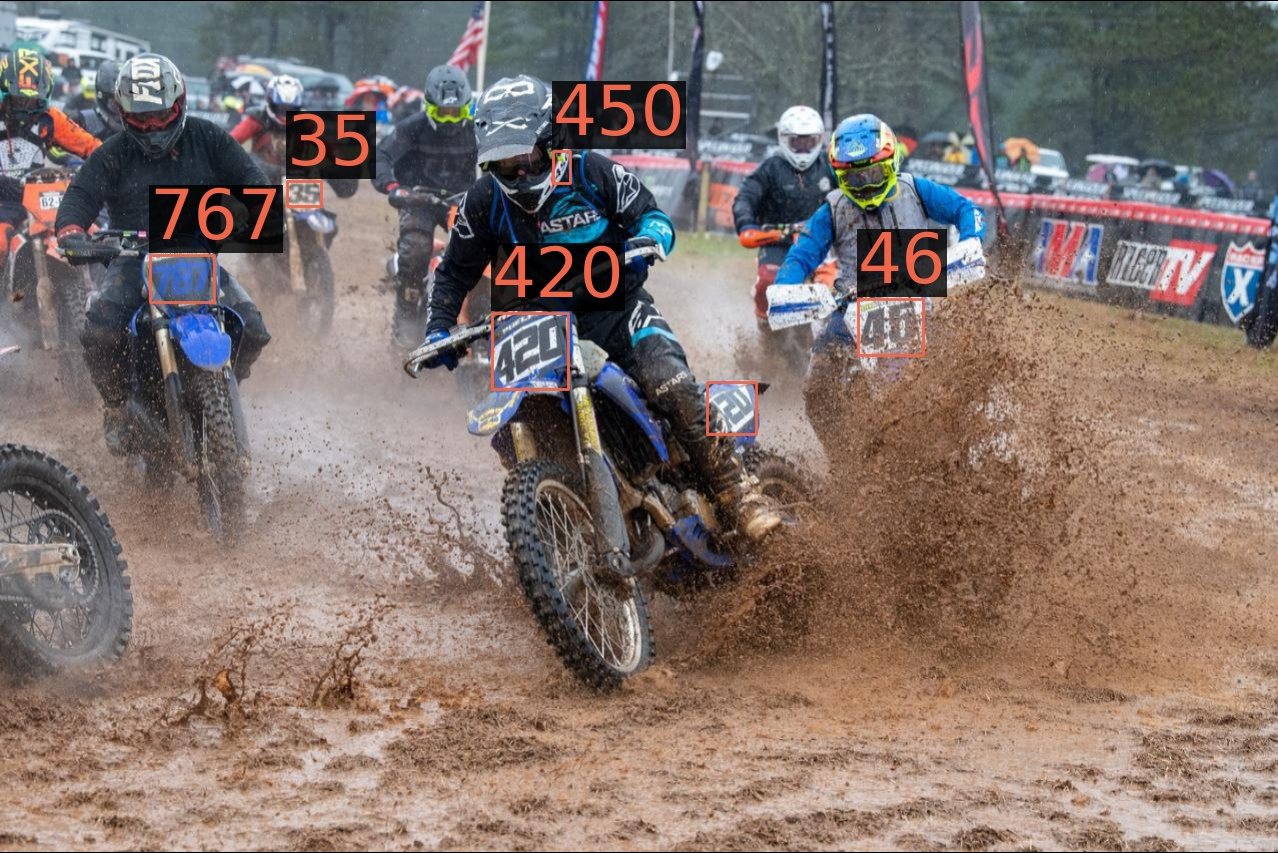}
    \end{subfigure}
    \caption{Example showcasing model improvement in rainy conditions. The top image shows detections from the off-the-shelf YAMTS model before fine-tuning, which recognizes only 1 number correctly (``35"). The bottom image displays results from the fine-tuned YAMTS model, which detects all 6 visible numbers and correctly recognizes 5 of them.}
    \label{fig:rainy-start}
\end{figure}

For both models, 
we first benchmark their performance on the RnD test set using 
their published pre-trained weights, which 
are from training on a large corpus of training data. 
YAMTS was pretrained on 
Open Images V5~\cite{OpenImages, krylov2021open}, 
ICDAR 2013~\cite{karatzas2013icdar}, 
ICDAR 2015~\cite{karatzas2015icdar}, 
ICDAR 2017~\cite{gomez2017icdar2017}, 
ICDAR 2019~\cite{zhang2019icdar}, 
COCO-text~\cite{veit2016coco}, 
and MSRA-TD500~\cite{yao2012detecting}. 
SwinTS was pretrained on 
Curved SynthText~\cite{liu2020abcnet},
TotalText~\cite{ch2017total}, 
ICDAR 2013~\cite{karatzas2013icdar}, 
and ICDAR-MLT~\cite{gomez2017icdar2017, zhang2019icdar}.

Afterwards, we fine-tune these models further
on the RnD training set and evaluate their performance again.
We first performed a grid search over the learning rate, learning rate schedule, warm-up period, and batch size using 
the validation set. 
We found the best setup to be a cosine annealing learning rate
schedule with a warm up, 
using a batch size of 8 images across 4 GPUs, 
with the random scaling and rotation data augmentations. 
The learning rate starts at 1e-6
and is then raised to 1e-3 after 1,000 iterations, 
and then annealed back down to 1e-6 over the 
remainder of training.
These hyperparameters were used to fine-tune the models 
over 150 epochs.
The fine-tuned models are evaluated on the RnD test set.

\subsection{Evaluation Metrics}

Following the standard evaluation protocol~\cite{huang2022swintextspotter, Ye_2023_CVPR}, 
we report results for both the text detection 
and end-to-end recognition tasks.
For detection, we compute precision, recall, and F1-score, which we denote \textit{Det-P}, \textit{Det-R}, and \textit{Det-F1} respectively. 
A predicted box was considered a true positive 
if it overlapped with a ground truth box 
by at least 50\% intersection over union.
For end-to-end recognition, 
we report precision, recall, and F1-score at the sequence level, and we likewise denote these metrics as \textit{E2E-P}, \textit{E2E-R}, and \textit{E2E-F1}. 
A predicted text sequence was considered correct 
only if it exactly matched the ground truth transcription for the corresponding ground truth box.

\section{Results and Discussion}

Table~\ref{tab:results} summarizes the quantitative results 
on the RnD test set. 
The off-the-shelf SwinTS and YAMTS models, 
which were 
pretrained on large generic OCR datasets, 
achieve poor accuracy. 
This highlights the substantial domain gap 
between existing datasets and this new motorsports application. 
Even state-of-the-art models fail 
without adaptation to racer numbers.

Fine-tuning the pretrained models on RnD 
led to major improvements. 
SwinTS achieved 0.734 detection F1 
and 0.459 end-to-end recognition F1 after fine-tuning. 
For YAMTS, 
fine-tuning improved to 0.775 detection 
and 0.527 recognition F1 scores. 
However, these fine-tuned results still 
fall short of requirements for robust real-world deployment.

The experiments reveal substantial room 
for improvement over state-of-the-art methods on RnD. 
Neither off-the-shelf nor fine-tuned models 
achieve sufficient accuracy for motorcycle racing applications, which we detail further in the next section with qualitative analysis. 
Overall, our quantitative benchmarks establish baseline results 
to motivate innovative techniques tailored 
to OCR on muddy vehicles in dynamic outdoor environments.

\subsection{Performance Among Occlusion}

\begin{table*}[]
\centering
\caption{Performance broken down by occlusion.}
\label{tab:occlusion_results}
\begin{tabular}{@{}lccccccc@{}}
\toprule
\multicolumn{2}{c}{Occlusion (\% of data)}    & Det-P & Det-R & Det-F1 & E2E-P & E2E-R & E2E-F1 \\ \midrule
\multirow{2}{*}{None (41\%)}  & Off-the-shelf & 0.196 & 0.568 & 0.291  & 0.124 & 0.330 & 0.180  \\
                              & Fine-Tuned    & 0.880 & 0.726 & 0.795  & 0.826 & 0.470 & 0.599  \\ \midrule
\multirow{2}{*}{Blur (3\%)}   & Off-the-shelf & 0.231 & 0.545 & 0.324  & 0.140 & 0.295 & 0.190  \\
                              & Fine-Tuned    & 0.860 & 0.841 & 0.851  & 0.750 & 0.409 & 0.529  \\ \midrule
\multirow{2}{*}{Shadow (7\%)} & Off-the-shelf & 0.144 & 0.536 & 0.227  & 0.033 & 0.107 & 0.050  \\
                              & Fine-Tuned    & 0.875 & 0.778 & 0.824  & 0.769 & 0.370 & 0.500  \\ \midrule
\multirow{2}{*}{Muddy (44\%)} & Off-the-shelf & 0.194 & 0.389 & 0.259  & 0.086 & 0.152 & 0.110  \\
                              & Fine-Tuned    & 0.811 & 0.718 & 0.761  & 0.681 & 0.359 & 0.470  \\ \midrule
\multirow{2}{*}{Glare (8\%)}  & Off-the-shelf & 0.162 & 0.547 & 0.250  & 0.052 & 0.156 & 0.078  \\
                              & Fine-Tuned    & 0.787 & 0.686 & 0.733  & 0.519 & 0.200 & 0.289  \\ \midrule
\multirow{2}{*}{Dust (2\%)}   & Off-the-shelf & 0.173 & 0.310 & 0.222  & 0.113 & 0.190 & 0.142  \\
                              & Fine-Tuned    & 0.925 & 0.638 & 0.755  & 0.833 & 0.259 & 0.395  \\ \bottomrule
\end{tabular}
\end{table*}

We further analyzed model performance on the RnD test set 
when numbers were occluded by different factors. 
Note that a single image can contain multiple occlusions (i.e. it can be dusty and have glare, or it can be blurry and muddy, etc.). 
Table \ref{tab:occlusion_results} breaks down the detection 
and recognition results on images with 
no occlusion, motion blur, shadows, mud, glare, and dust.

Mud occlusion was the most prevalent, accounting for 44\% 
of the test data. 
Both off-the-shelf and fine-tuned models 
struggled with heavy mud. 
The fine-tuned model improved over the off-the-shelf version, 
achieving 0.761 detection F1 and 0.470 recognition F1 
on muddy images. 
But this remains far below the 0.795 detection 
and 0.599 recognition scores attained 
on non-occluded data. 
There is substantial room to improve 
robustness to real-world mud and dirt occlusion.

The fine-tuned model also struggled with glare occlusion, 
scoring just 0.733 detection F1 
and 0.289 recognition F1 on such images. 
Glare creates low contrast regions that 
likely hurt feature extraction. 
Shadows likewise proved challenging, 
with a 0.824 detection but 
only 0.500 recognition F1 score after fine-tuning. 
The changing lighting and hues may degrade recognition.

For motion blur, the fine-tuned model achieved 0.851 detection F1 
but 0.529 recognition F1. 
Blurring degrades the crispness of text features needed 
for accurate recognition. 
Surprisingly, the model performed worst on dust occlusion, 
despite it being visually less severe than mud and glare. 
This highlights brittleness of vision models to unusual textures.

Overall, 
the breakdown reveals mud as the primary challenge, 
but substantial room remains to improve OCR accuracy 
under real-world conditions 
like shadows, dust, blur, and glare. 
Researchers should prioritize occlusions seen 
in natural operating environments 
that undermine off-the-shelf models.

\subsection{Qualitative Analysis}

We analyzed model performance on RnD using 
the fine-tuned YAMTS model, 
which achieved the highest end-to-end F1 score. 
The detection confidence threshold was set to 0.65 
and the recognition threshold set to 0.45. 
Figures~\ref{fig:muddy-start}-\ref{fig:rainy-start} 
showcase successes and failures on challenging examples.
When side-by-side comparisons are drawn, 
we compare against the off-the-shelf YAMTS model 
before fine-tuning. 

Figure~\ref{fig:muddy-start} 
compares the text spotting performance before 
and after fine-tuning on a photo 
of the start of a muddy race. 
The fine-tuned model properly detects 
all 8 visible numbers, 
demonstrating capabilities to handle partial mud occlusion. 
However, 
it only correctly recognizes 3 of the 8 numbers, 
highlighting limitations recognizing degraded text. 
Without fine-tuning, 
only 1 number is detected, 
and no numbers are properly recognized, 
showing benefits of fine-tuning.
But substantial challenges remain in muddy conditions.

Figure~\ref{fig:mudfails} showcase common mud-related successes and failures. 
In some casese, 
the fine-tuned models are able to see through mud occlusions to properly recognize 
the racer number, as shown in Figure~\ref{img:rtm2}.
However, mud often prevents smaller helmet numbers 
from being recognized (Fig~\ref{img:rtm1}, \ref{img:mudfail}). 
Odd orientations also confuse models (Fig~\ref{img:mudslidefail}). 
Overall, heavy mud occlusion remains the biggest challenge.
Figure~\ref{fig:cleanfails} reveals other common failures 
like missing side numbers (Fig~\ref{img:backbikefail}), 
overlapping numbers (Fig~\ref{img:stackednums}), 
confusion between letters and numbers (Fig~\ref{img:wrongletter}), 
missing letter portions (Fig~\ref{img:missedletter}), 
and distractions from graphics (Fig~\ref{img:quadshard}).
In summary, 
the analysis reveals promising capabilities 
but also exposes key areas for improvement, 
particularly among extreme mud and small text.
Substantial opportunities remain to enhance OCR 
for this challenging real-world application.

Photos from the beginning of a race are typically the most complex,
due to the number of motorcycles in a single image and background clutter. 
Figure~\ref{fig:rainy-start} again looks at a photo from the 
start of a race, but this time in rainy conditions. 
The top photo highlights the detections of the off-the-shelf model before
fine-tuning, where it is able to recognize only a single number properly. 
However, after fine-tuning, 
the model is able to properly recognize 5 of the 6 visible numbers.

\section{Conclusion} 

In this work, 
we introduced the 
off-road motorcycle Racer number Dataset (RnD), 
a novel challenging real-world dataset 
to drive advances in optical character recognition. 
RnD contains 2,411 images exhibiting factors such as 
mud, motion blur, glare, complex backgrounds, and occlusions 
that degrade text detection and recognition accuracy. 
The images were captured by 
professional motorsports photographers across 
50 distinct off-road competitions.

We annotated 
5,578 racer numbers 
with transcriptions and tight bounding boxes. 
The data exhibits natural diversity 
in lighting, weather, track conditions, 
vehicle types, racer gear, and more. 
To our knowledge, 
RnD represents the largest, 
most varied collection of annotated motorsports numbers 
in unconstrained environments.

We established baseline results on RnD 
using the state-of-the-art text spotting models, 
Swin Text Spotter and YAMTS. 
Off-the-shelf versions pretrained 
on generic OCR data achieved an end-to-end F1 score 
around 0.2, highlighting the sizable domain gap. 
Fine-tuning on RnD improved results 
but even the best model obtained only 0.527 end-to-end F1, 
far below practical expectations for real-world use.
Through qualitative analysis, 
we revealed some of the primary factors 
degrading OCR accuracy on RnD 
to be heavy mud occlusion, glare, dust, and more. 
Heavily distorted fonts and unusual orientations 
also led to several notable mistakes. 

Overall, 
our work exposes motorcycle racer number recognition 
as an open challenge with unique conditions, 
and provides a dataset of novel real-world imagery. 
The experiments establish baseline results using leading methods, 
quantitatively and quantitatively 
demonstrating substantial room for improvement on RnD. 
We hope the community will build upon these initial experiments 
to make advances on the problem of accurately 
reading text in unconstrained natural environments.

{\small
\bibliographystyle{ieee_fullname}
\bibliography{references}
}

\end{document}